\documentclass[conference]{IEEEtran}
\IEEEoverridecommandlockouts
\usepackage{pgfplots}
\usetikzlibrary{patterns}
\usepackage{stfloats}
\usepackage{placeins}
\usepackage{array}
\usepackage{calc}
\usepackage{array, booktabs, makecell}
\usepackage{amsmath,amssymb,amsfonts}
\usepackage{algorithmic}
\usepackage[caption=false]{subfig}
\usepackage{tikz}
\usepackage{graphicx}
\usepackage{textcomp}
\usepackage{cite}

\usepackage{hyperref}
\usepackage{cleveref}
\crefname{figure}{Figure}{Figures}
\crefname{table}{Table}{Tables}
\crefname{algorithm}{Algorithm}{Algorithms}
\crefname{equation}{Equation}{Equations}
\crefname{section}{Section}{Sections}

\usepackage{algorithm}
\usepackage{xcolor}

\pgfplotsset{compat=newest,width=8cm,height=4.8cm,
        every axis/.append style={
            tick label style={font=\fontsize{6}{6.5}\selectfont},
            label style={font=\fontsize{6}{6.5}\selectfont}
        },
        legend image code/.code={
            \draw[mark repeat=2,mark phase=2]
                plot coordinates {
                    (0cm,0cm)
                    (0.15cm,0cm)        
                    (0.3cm,0cm)
                };
        },
        major tick length=0.03cm,
        xtick align=outside,ytick align=outside,
        axis x line*=bottom,axis y line*=left,axis line style=ultra thin
}

\makeatletter % changes the catcode of @ to 11
\newcommand{\linebreakand}{%
  \end{@IEEEauthorhalign}
  \hfill\mbox{}\par
  \mbox{}\hfill\begin{@IEEEauthorhalign}
}
\makeatother % changes the catcode of @ back to 12

\def\BibTeX{{\rm B\kern-.05em{\sc i\kern-.025em b}\kern-.08em
    T\kern-.1667em\lower.7ex\hbox{E}\kern-.125emX}}
\begin{document}

\title{INGREX: An Interactive Explanation Framework for Graph Neural Networks}

\author{\IEEEauthorblockN{1\textsuperscript{st} Tien-Cuong Bui}
\IEEEauthorblockA{\textit{Department of ECE} \\
\textit{Seoul National University}\\
Seoul, South Korea \\
cuongbt91@snu.ac.kr}
\and
\IEEEauthorblockN{2\textsuperscript{nd} Van-Duc Le}
\IEEEauthorblockA{\textit{Department of ECE} \\
\textit{Seoul National University}\\
Seoul, South Korea \\
levanduc@snu.ac.kr}
\and
\IEEEauthorblockN{3\textsuperscript{rd} Wen-Syan Li}
\IEEEauthorblockA{\textit{Graduate School of Data Science} \\
\textit{Seoul National University}\\
Seoul, South Korea \\
wensyanli@snu.ac.kr}
\and
% \linebreakand
\IEEEauthorblockN{4\textsuperscript{th} Sang Kyun Cha}
\IEEEauthorblockA{\textit{Graduate School of Data Science} \\
\textit{Seoul National University}\\
Seoul, South Korea \\
chask@snu.ac.kr}

}

\maketitle

\begin{abstract}
Graph Neural Networks (GNNs) are widely used in many modern applications, necessitating explanations for their decisions. However, the complexity of GNNs makes it difficult to explain predictions. Even though several methods have been proposed lately, they can only provide simple and static explanations, which are difficult for users to understand in many scenarios. Therefore, we introduce INGREX, an interactive explanation framework for GNNs designed to aid users in comprehending model predictions. Our framework is implemented based on multiple explanation algorithms and advanced libraries. We demonstrate our framework in three scenarios covering common demands for GNN explanations to present its effectiveness and helpfulness.
\end{abstract}


\section{Introduction} 
% the need for GNN explanations, the need for visualization of GNN explanations
% challenges or interesting points

The widespread use of Graph Neural Networks (GNNs) \cite{zhou2020graph} has resulted in the need to explain their predictions, especially when making critical decisions. However, the complexity of graph data and model execution makes it difficult to explain GNNs' predictions. A certain prediction's influential factors may come from complicated interactions among elements within an input graph. Therefore, it is necessary to measure and visualize the contributions of these interactions in a way that users can understand.

Since explainable AI (XAI) has recently been a hot research topic, many frameworks and methods have been proposed. InterpretML \cite{nori2019interpretml} is a unified framework that provides various functions for interactive explanations of ML models. However, it does not support GNN models. SHAP \cite{lundberg2017unified} and LIME \cite{ribeiro2016should} are famous feature attribution frameworks that provide explanation features with intuitive visualization for different models. However, they do not support GNN models and interactive interpretation. Even though several GNN explanation methods \cite{yuan2020explainability} have been proposed, their visualizations are usually static and simple, sometimes leaving users and practitioners with less intuition. 

% what do we do
We address these issues in INGREX, an \textbf{IN}teractive \textbf{GR}aph Neural Network \textbf{EX}planation framework. Our framework exposes GNN explanation methods under interactive features and aims to increase users’ comprehension of model predictions. Specifically, it consists of two components frontend and backend. The frontend is responsible for receiving user requests and visualizing explanation results. It supports interactive features via a web application based on Angular and CytoscapeJS. For instance, a global visualization of a node classification graph provides users with information such as decision boundaries. After the comprehensive examination, users can view the local explanations of target nodes. They understand predictions better in graph classification when comparing the explanation of the target graph with ones of references from different classes. Based on algorithms proposed in \cite{pgxalex,scalealex}, the backend implements multiple functions specialized for different scenarios. Our framework is open to integrating additional explanation methods since it is flexible and extensible. Once users require explanations for predictions, the web application sends REST requests to the backend and visualizes the results with interactive features. We demonstrate our framework’s helpfulness and effectiveness in three scenarios representing popular requirements for GNN explanations.

\section{Technical Background} \label{tech_background}
\subsection{Graph Neural Networks}
GNNs mostly compute node embeddings via iteratively executing propagation, aggregation, and update operations. Let $h^{l}_i$ be the representation vector $i^{th}$ at the layer $l^{th}$. A message $m^l_{ij} = \textrm{Message}(h^{l-1}_i, h^{l-1}_j)$ is sent along an edge $ji$.
Then, information is aggregated by $m^l_i = \textrm{Aggregate}({m^l_{ij}|j \in \mathcal{N}_i})$. A representation vector is updated by $h^l_i = \textrm{Update}(m^l_i, h^{l-1}_i)$. The embedding matrix $h^L$ of the last layer can be the input of downstream models such as node and graph classification.

\subsection{Constructing Explanations for GNNs}
We must find influential factors of a GNN prediction from an input graph $G_c$. These factors are complicated interactions among the graph structure, node features, and edge features. Edge features are usually discarded for simplicity. An explanation is formulated as ($G_s, \Phi_x$), where $G_s$ is a subgraph of $G_c$ and $\Phi_x$ denotes node feature attributions. In the next two subsections, we summarize explanation methods for GNNs proposed in \cite{pgxalex,scalealex}.

\subsection{Explaining Node-level Predictions}
\noindent\textbf{Structural Explanation.} Bui et al. \cite{pgxalex,scalealex} propose employing random walk with restart (RWR) to construct structural explanations for node-level predictions. Let us first recall the RWR formula:
\begin{equation}
    r_{t+1} = (1 - d)r_0 + d \hat{\mathcal{A}}_c r_t,
    \label{eq:rwr_alg}
\end{equation}
where $\hat{\mathcal{A}}_c$ is a column-normalized matrix, $r_t$ denotes the probability distribution at time $t$, $r_0$ is the initial probability distribution, and $d$ is the keep-going probability. Practically, $\mathcal{A}_c$ is a transposed version of a row-normalized adjacency matrix, which can be constructed in many ways. For instance, one can normalize attention heads of GAT \cite{velickovic2017graph,pgxalex,scalealex} or replace the symmetrically normalized adjacency matrix with a trainable matrix \cite{scalealex}. A target node is considered a query node, wherein only its corresponding element in $r_0$ has a non-zero value. A subgraph containing top-$k$ nodes is a structural explanation of a node-level prediction.

\noindent\textbf{Measuring Feature Attributions.} Bui et al. \cite{pgxalex,scalealex} propose transforming the GNN model into smaller, easier-to-explain models using knowledge distillation (KD) paradigms. A multilayer perceptron (MLP) model is trained to imitate the behaviors of the GNN model. Then, attribution methods such as \cite{lundberg2017unified} are executed on the MLP model to examine feature attributions.

\subsection{Explaining Graph-level Predictions}
\noindent\textbf{Structural Explanation.} Based on \cite{scalealex}, we describe the structural explanation procedure of a GCN model \cite{kipf2016semi} for simplicity. Here is the simplest layer-wise propagation operation:

\begin{equation}
    f(H^l, A) = \sigma(A H^l W^l),
    \label{basic_gcn}
\end{equation}
where $A$ denotes the adjacency matrix, $H^l$ is a representation matrix, $W^l$ is a weight matrix, and $\sigma$ is a non-linear function. Next, a mask matrix $M$ is added to \cref{basic_gcn} as follows:
\begin{equation}
    f(H^l, A, M) = \sigma((A \odot M)H^l W^l).
    \label{eq:graph_mask}
\end{equation}
$M$ is initialized via an MLP model as follows:
\begin{equation}
    m_{ij} = sigmoid(\textrm{MLP}([h_i, h_j])),
\end{equation}
\noindent where $h_i$ and $h_j$ are embeddings of two nodes of an edge. The self-explainable GCN model is trained simultaneously with an original GCN model based on a KD paradigm \cite{pgxalex,scalealex}. In the inference phase, $M$ provides structural explanations of graph-level predictions.

\noindent\textbf{Example-based Explanation.} This function provides explanations of references from different classes for comparison. This approach is well-studied in XAI for computer vision and NLP problems \cite{jeyakumar2020can}. However, it is less explored in GNN explanation methods. To our knowledge, no example-based explanation methods for GNNs are currently available. Combined with structural explanations, this method gives users more insights into model predictions.

\section{System Architecture}
% what is our system architecture?
\subsection{System Description}
\begin{figure}
    \centering
    \includegraphics[width=\columnwidth,height=5.9cm,trim={3.8cm 0cm 1.6cm 0.2cm},clip]{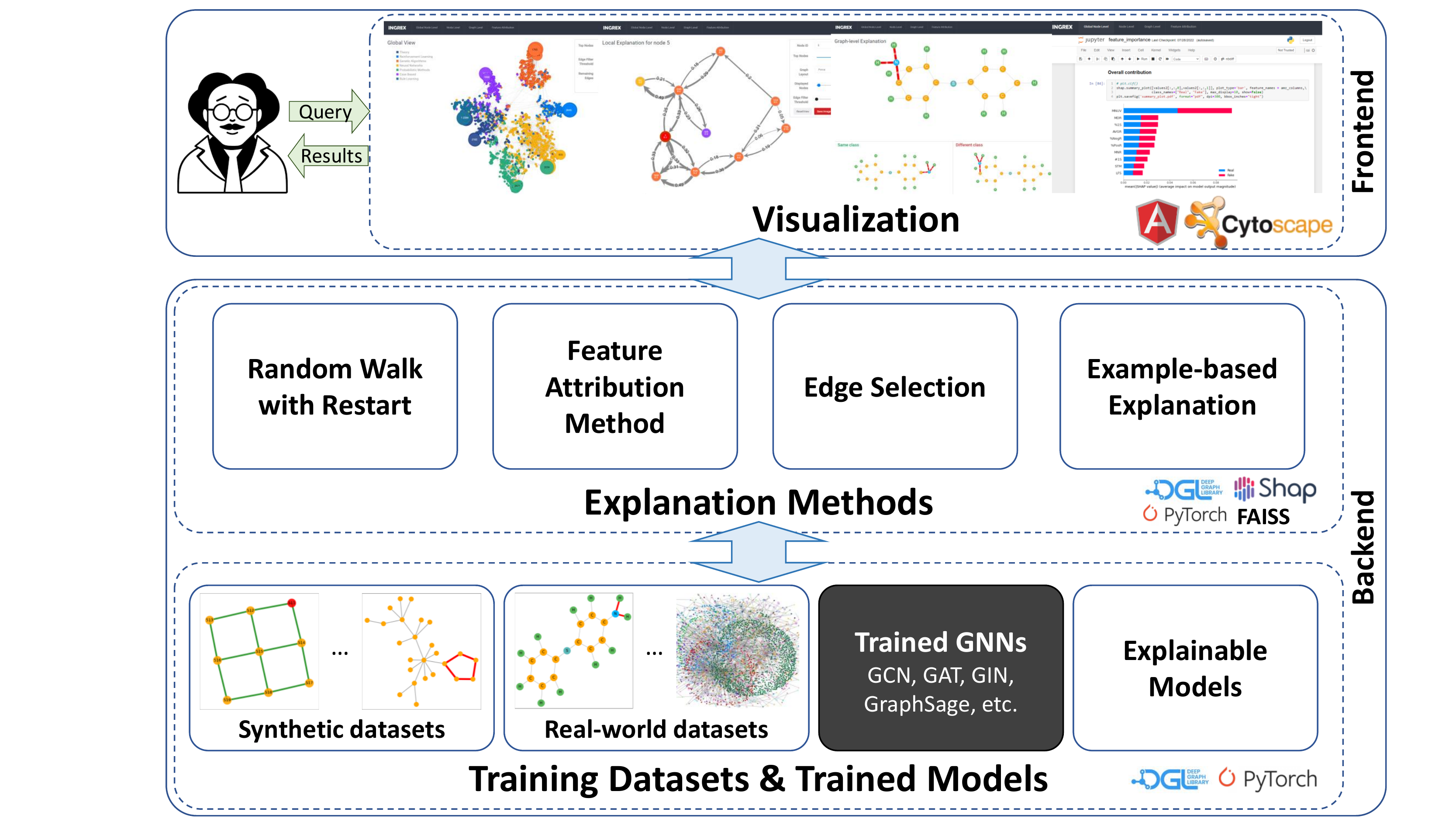}
    \caption{The INGREX System Architecture}
    \label{fig:my_label}
\end{figure}

\noindent\textbf{Backend.} We implement a Python web server using Flask. The backend consists of two main components: trained models and explanation methods. Currently, our system manages datasets and models as files and loads them into memory when needed. Explanation methods correspond to algorithms described in \cref{tech_background}. The Python web server allows us to integrate several libraries using for model execution and explanation methods, such as DGL\cite{dgldata}, PyTorch, NetworkX, SHAP\cite{shap_doc}, and Faiss \cite{johnson2019billion}. DGL and Pytorch are used for module training and inference. SHAP provides feature attributions of predictions. Faiss, a billion-scale similarity search, is utilized for providing example-based explanations.

\noindent\textbf{Frontend.} We implement a web application to receive user requests and visualize explanations using Angular and CytoscapeJS. We select Angular since it is an efficient framework for building cross-platform web applications and has a large developer community. CytoscapeJS \cite{Cytoscape} is a Javascript library of Cytoscape, a powerful open-source platform for visualizing complex networks and providing useful UI/UX features. Similar to Angular, CytoscapeJS allows additional plugins to be integrated. Visualization Examples are presented in \cref{fig:example_demo}.

\noindent\textbf{Users.} Explanations provided by our system are the best fit for ML practitioners, model developers, and domain experts. ML practitioners and model developers can use explanation results to detect the abnormalities of datasets and trained models. By that, they can improve the predictive accuracy of models. On the other hand, domain experts can get insights from explanations or confirm results with their hypotheses. 

\noindent\textbf{Operational Flows.} When a user requests an explanation for a particular prediction, the frontend web application sends a REST request to the backend server. The backend server parses the request and executes a corresponding explanation method based on an explainable model and a training dataset. After that, it returns an explanation result to the web application to show to the user. After being loaded into the memory, models and datasets are retained to reduce the execution time of future requests. 

\subsection{Improvements and Extensions}
Since our system is currently a prototype version, there is plenty of room for improvements and extensions. First, in-memory libraries can be integrated into our system to optimize model and data management. Second, caching and load-balancing techniques can accelerate the system's performance further. Third, our system is open to integrating additional explanation methods. Fourth, we can design a text generation module on top of explanation methods to personalize explanations for various groups of users. Finally, combining the results of different explanation methods into one view further facilitates user predictions' comprehension.

\section{Demonstration Scenarios}

\begin{figure*}[ht]
\centering
    \subfloat[Global View of a Graph]{
        \includegraphics[width=0.32\linewidth,height=4cm,trim={0 1.5cm 0 0.1cm},clip]{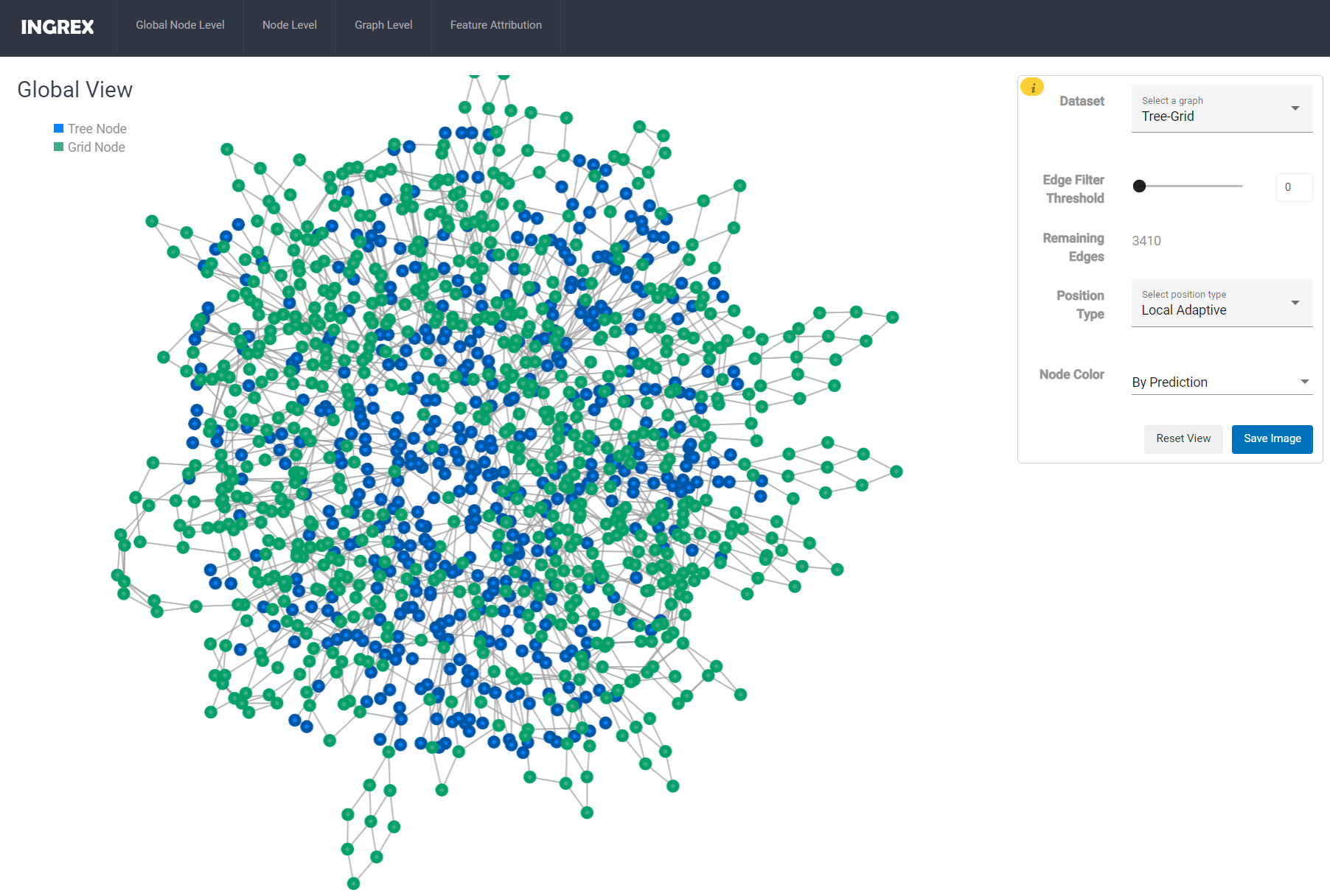}
    }
    \subfloat[Explanation of a Node Prediction]{
        \includegraphics[width=0.32\linewidth,height=4cm,trim={0 1.5cm 0 0.1cm},clip]{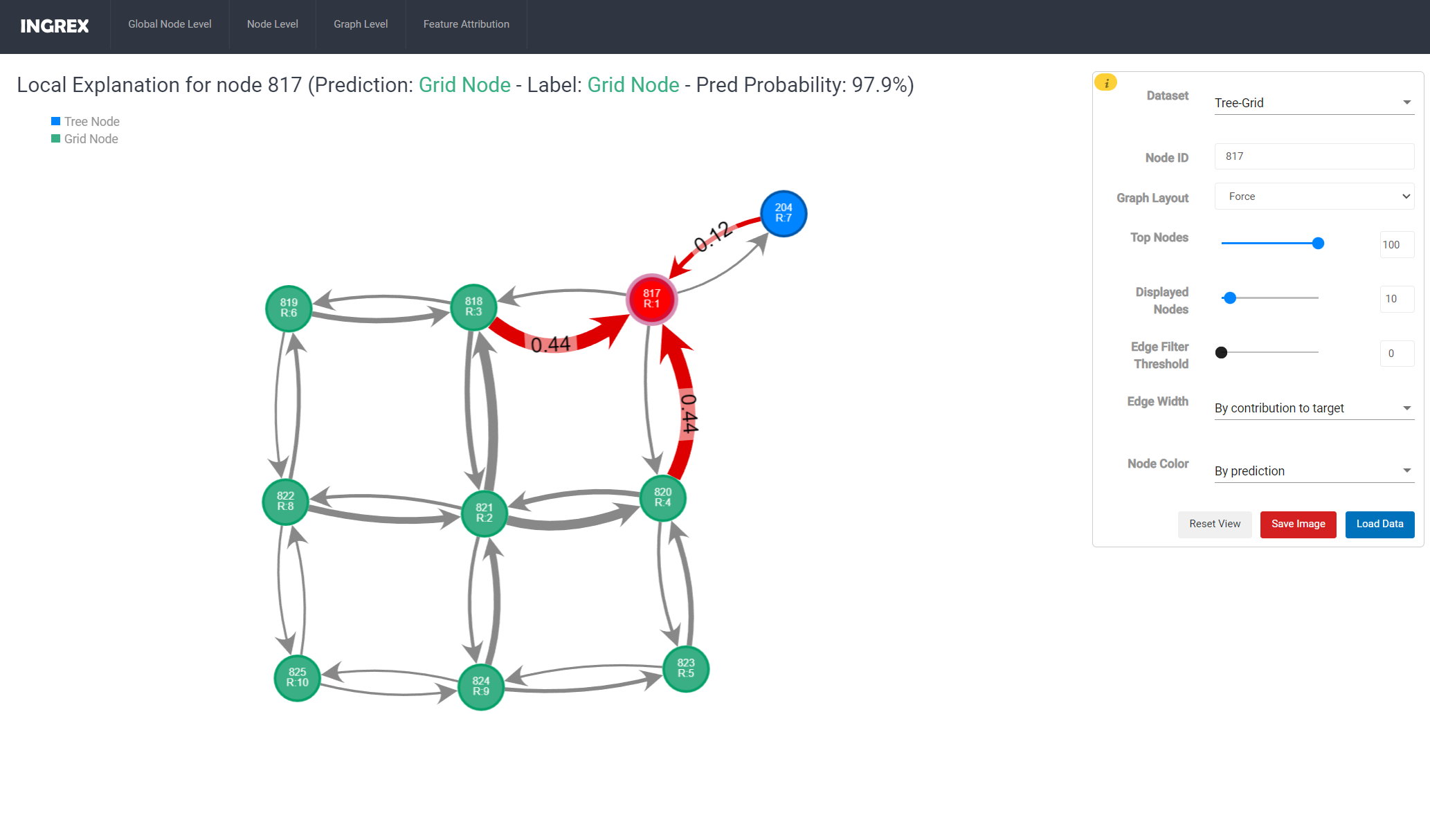}
        \label{fig:node_ex}
    }
    \subfloat[Explanation of a Graph Prediction]{
        \includegraphics[width=0.32\linewidth,height=4cm,trim={0 1.5cm 0 0.1cm},clip]{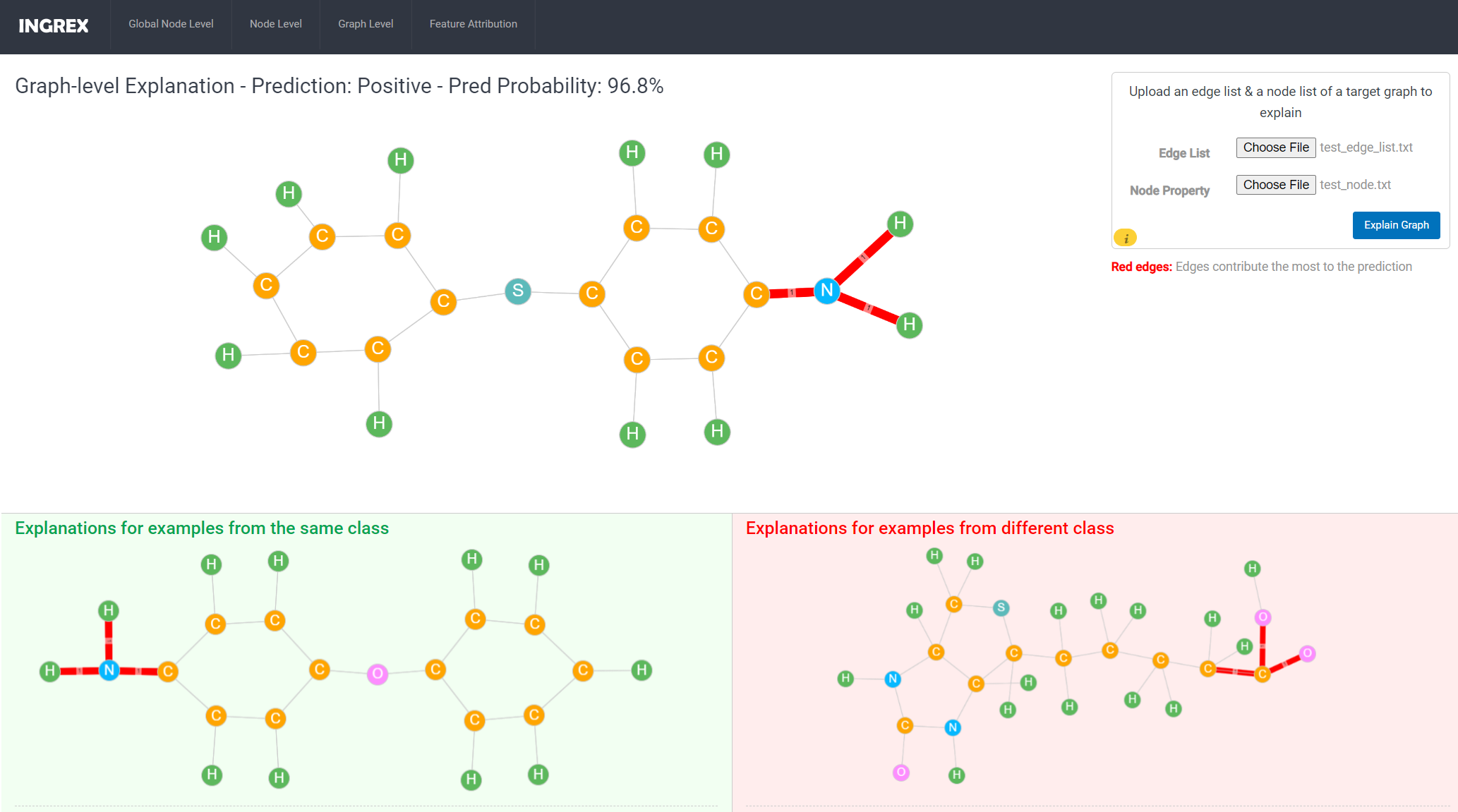}
        \label{fig:graph_ex_wd}
    }
    
    \caption{Visualization of Explanations on the Web Application}
    \label{fig:example_demo}
\end{figure*}

\subsection{Objectives}
This demonstration illustrates the capability of INGREX on interactive explanations of GNNs' predictions. We aim to show that there is no silver bullet for XAI in GNNs. Therefore, multiple explanation methods allow users to confirm prior hypotheses and extract insights from explanations. Additionally, the demo highlights the potential and limitations of the current framework version. 

The demonstration includes three scenarios corresponding to explaining node-level and graph-level predictions. First, we present multilevel structural explanations of node predictions. Next, we present how INGREX supports feature attribution measurement of node predictions. The last one illustrates structural and example-based explanations for graph-level predictions. The source code is available at \href{https://github.com/alexbui91/INGREX}{https://bit.ly/INGREX}.

\subsection{Datasets}
This demonstration used five datasets Tree-Grid, Cora, Amazon, BA-2motifs, and Mutag. The first is a synthetic node classification dataset introduced in \cite{ying2019gnnexplainer}. Cora is a popular benchmark dataset \cite{dgldata} for node classification. Next, Amazon \cite{dgldata,zhang2020gcn} is prevalent in fraudulent user detection research. The last two datasets are widely used to benchmark explanation methods \cite{ying2019gnnexplainer,luo2020parameterized} for graph classification.

\subsection{Scenario 1: Structural Explanations of Node Classification}
\noindent\textbf{Input.} An input graph $G = (V, E, X)$ consists of a set of nodes $V$, a set of edges $E$, and a node feature matrix $X$. The goal is to find crucial neighbors and edges that contribute the most to a certain node-level prediction.

\noindent\textbf{Process.} INGREX first visualizes the input graph based on either node embeddings or a local adaptation method. Users can select target nodes from the visualization to perform local explanations. The backend parses an explanation request to a corresponding graph dataset and a target node. Based on a self-explainable GNN trained by \cite{pgxalex,scalealex}, \cref{eq:rwr_alg} is executed on the column-normalized matrix $\mathcal{A}_c$ to provide explanations. After that, the backend returns a subgraph containing contributions of edges to the target node.

\begin{figure}[ht]
    \centering
    \subfloat{
        \includegraphics[width=0.7\columnwidth,trim={0.7cm 15.7cm 0.7cm 1.8cm},clip]{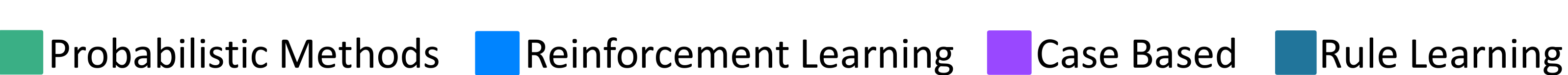}
    }
    \setcounter{subfigure}{0}
    \hfil
    \subfloat[True Prediction]{
        \includegraphics[width=0.48\columnwidth,trim={0.3cm 0.3cm 0.3cm 0.2cm},clip]{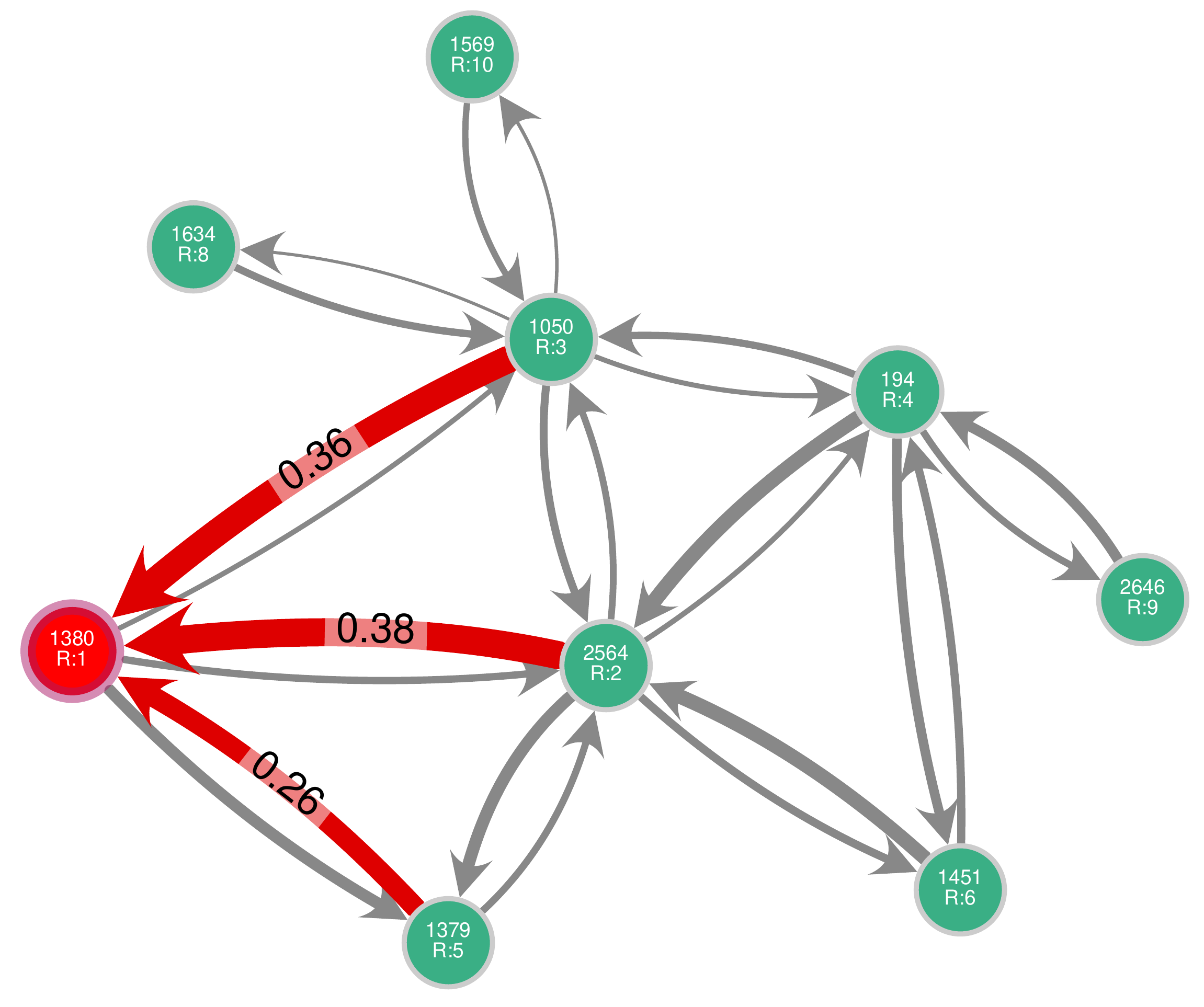}
    }
    \subfloat[Wrong Prediction]{
        \includegraphics[width=0.48\columnwidth,trim={0.4cm 0.4cm 0.4cm 0.6cm},clip]{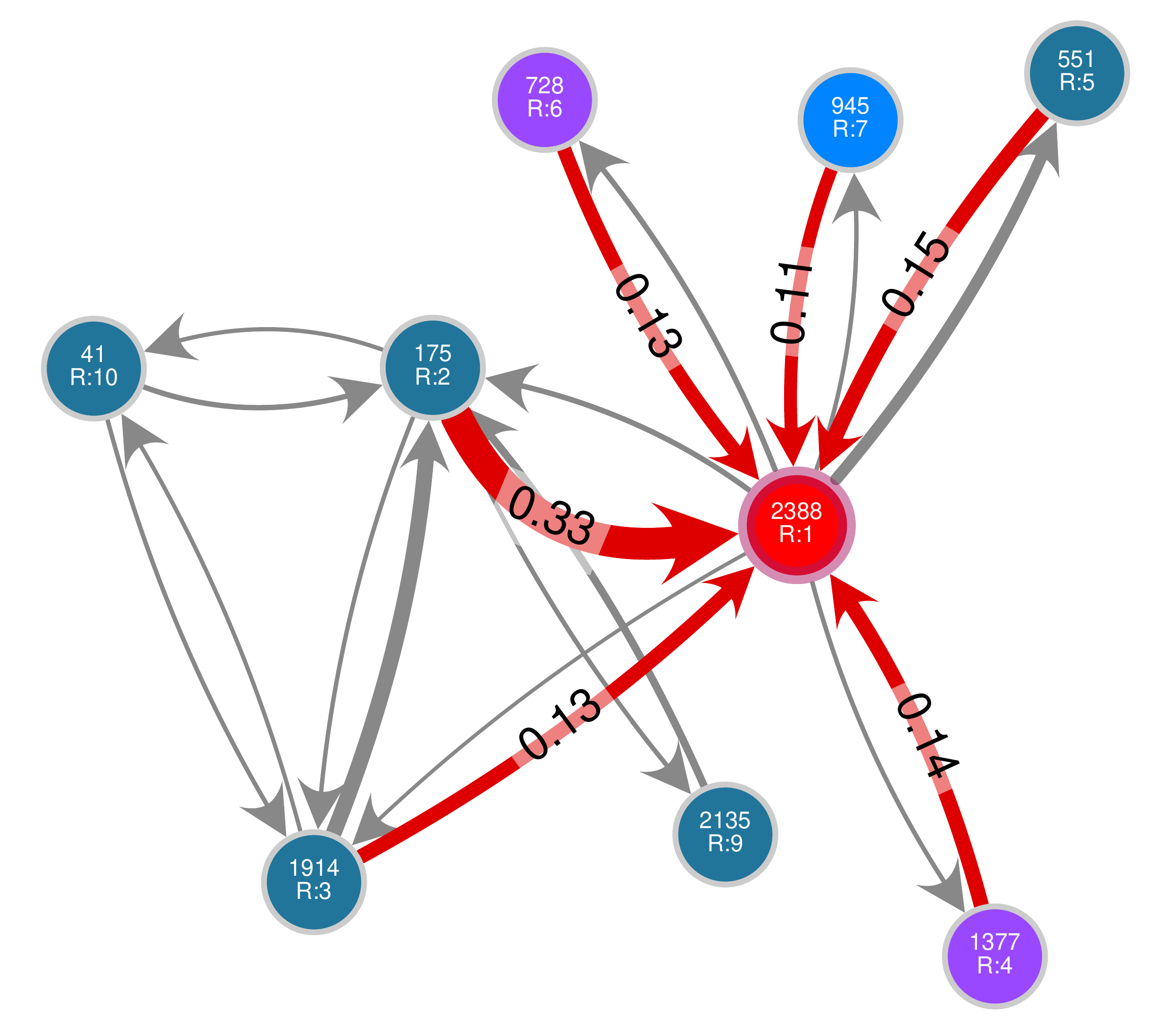}
    }
    \caption{Structural Explanations of Node-level Predictions in Cora Dataset. Node colors illustrate different classes. The red node is the explained one. Red edges highlight contributions of 1-hop neighbors with corresponding ratios. Contributions can be viewed on multiple levels (k-hop neighbors).}
    \label{fig:node_structure}
\end{figure}

\noindent\textbf{Output.} Example explanations are presented in \cref{fig:node_ex,fig:node_structure}. Based on explanations, users can see the contributions of k-hop neighbors to a target node prediction. They can also understand the reasons for wrong predictions when nodes locate in the boundary among classes and are impacted by cross-class edges.

\subsection{Scenario 2: Feature Attributions of Node Classification}
\noindent\textbf{Input.} The goal is to find the overall contributions of node features to model predictions and specific feature attributions of a certain node-level prediction. The Amazon dataset is used in this scenario.

\noindent\textbf{Process.} Here, we integrated Jupyter into our web application allowing users to interact directly with feature attribution methods. Since numerous attribution frameworks have been proposed \cite{shap_doc,nori2019interpretml}, the integration enables users to select a method that satisfies their demands freely. For simplicity, we tried several methods provided by SHAP \cite{shap_doc} on an MLP model guided by a GNN model in training. 

\begin{figure}[ht]
    \centering
    \includegraphics[width=0.98\columnwidth]{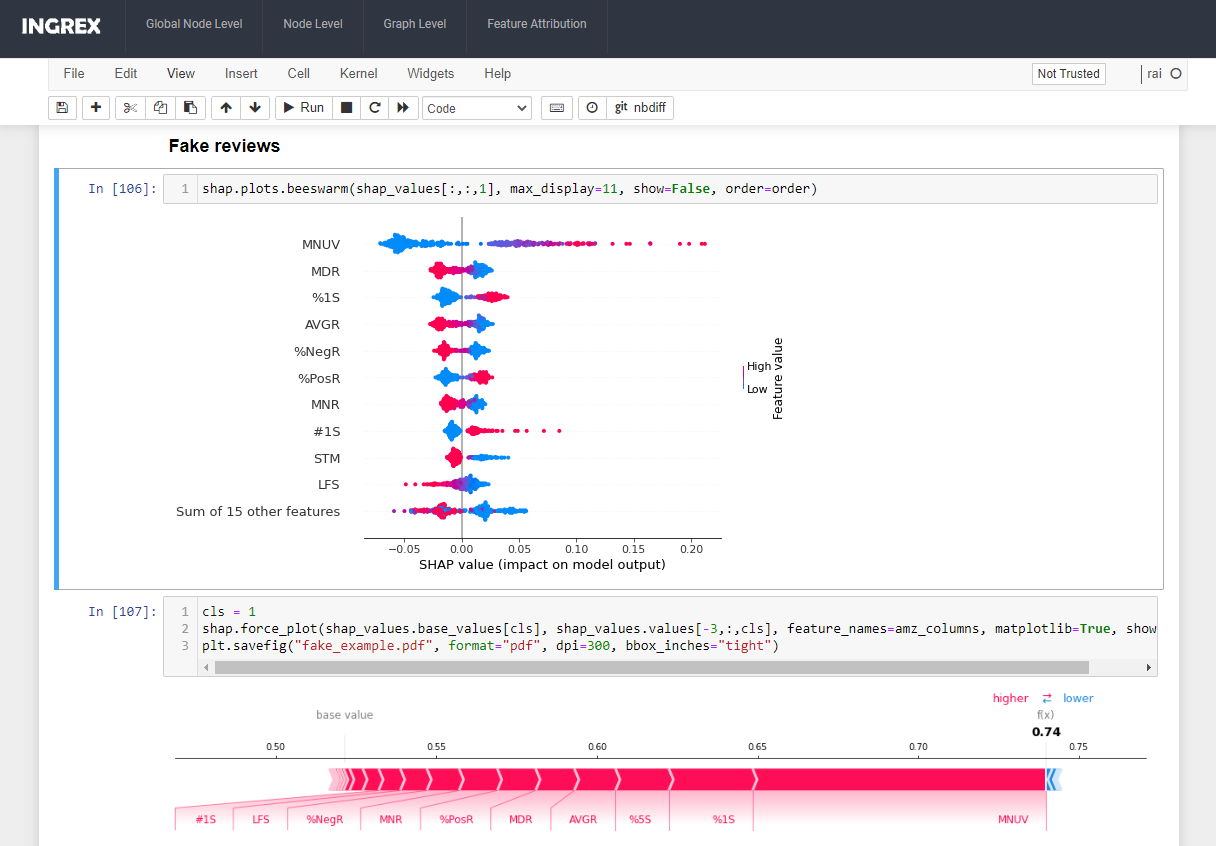}
    \caption{Visualization of Feature Attributions in INGREX}
    \label{fig:instance_explanations}
\end{figure}

\noindent\textbf{Output.} Explanation examples are depicted in \cref{fig:instance_explanations}, including a contribution summary and an instance explanation. Global summarizations allow users to understand overall feature contributions, while local attributions clarify the exact feature influences on particular predictions.

\subsection{Scenario 3: Explaining Graph-level Predictions}
\noindent\textbf{Input.} Given an input graph similar to samples in training datasets, the goal is to find crucial motifs in the input graph that leads to a certain prediction. We used BA-2motifs and Mutag in this scenario.

\noindent\textbf{Process.} The backend executes an edge selection procedure on the mask matrix $M$ of a pre-trained self-explainable GNN \cite{scalealex}. Next, the system returns the structural explanation with an example-based explanation, which includes explanations of samples from the same and different classes.

\begin{figure}[ht]
    \centering
    \hfil
    \subfloat[Target Instance]{
        \includegraphics[width=0.31\columnwidth,trim={2.5cm 2.5cm 2cm 2.4cm},clip]{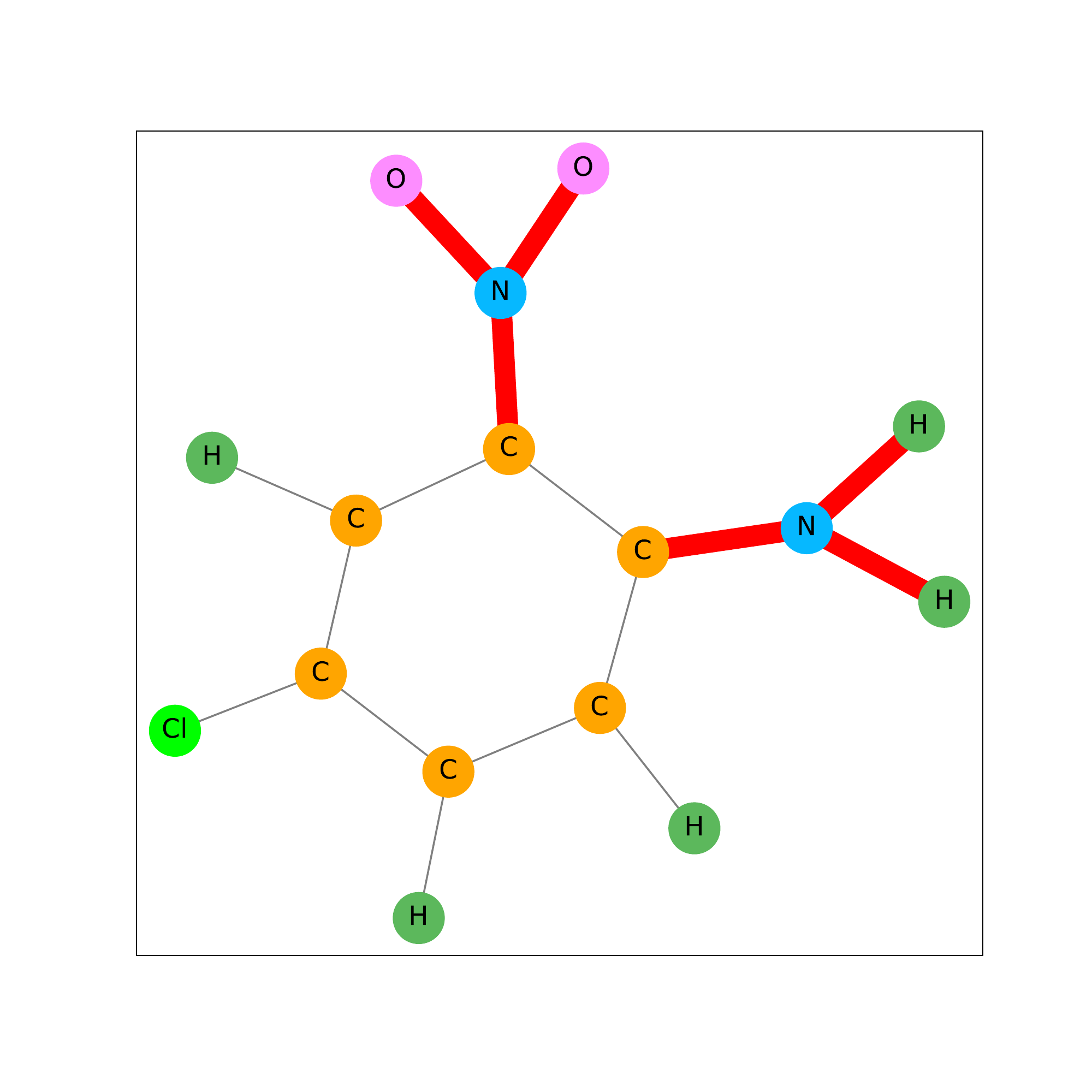}
    }
    \subfloat[Same Class]{
        \includegraphics[width=0.31\columnwidth,trim={2.5cm 2.5cm 2cm 2.4cm},clip]{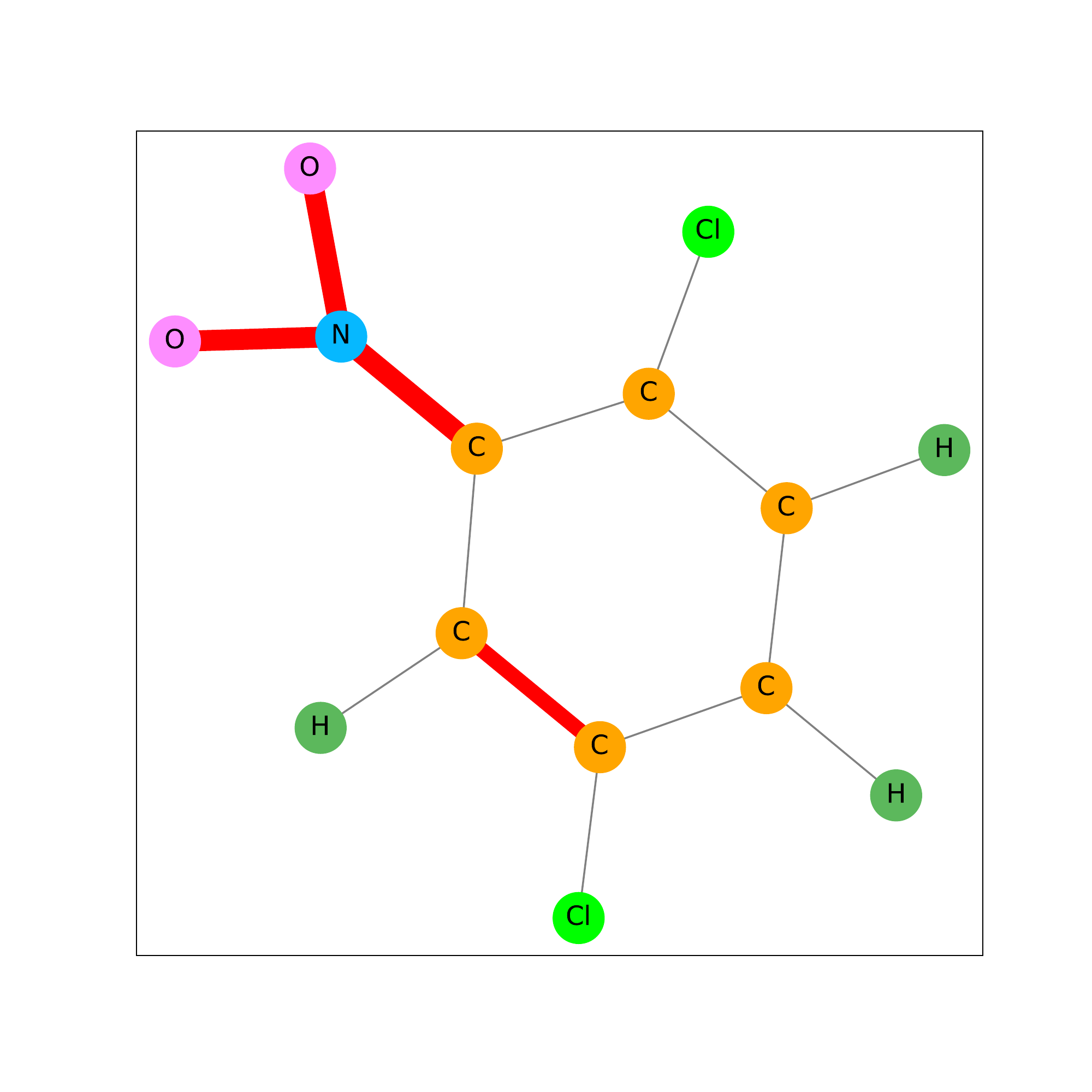}
    }
    \subfloat[Different Class]{
        \includegraphics[width=0.31\columnwidth,trim={2.5cm 2.5cm 2cm 2.4cm},clip]{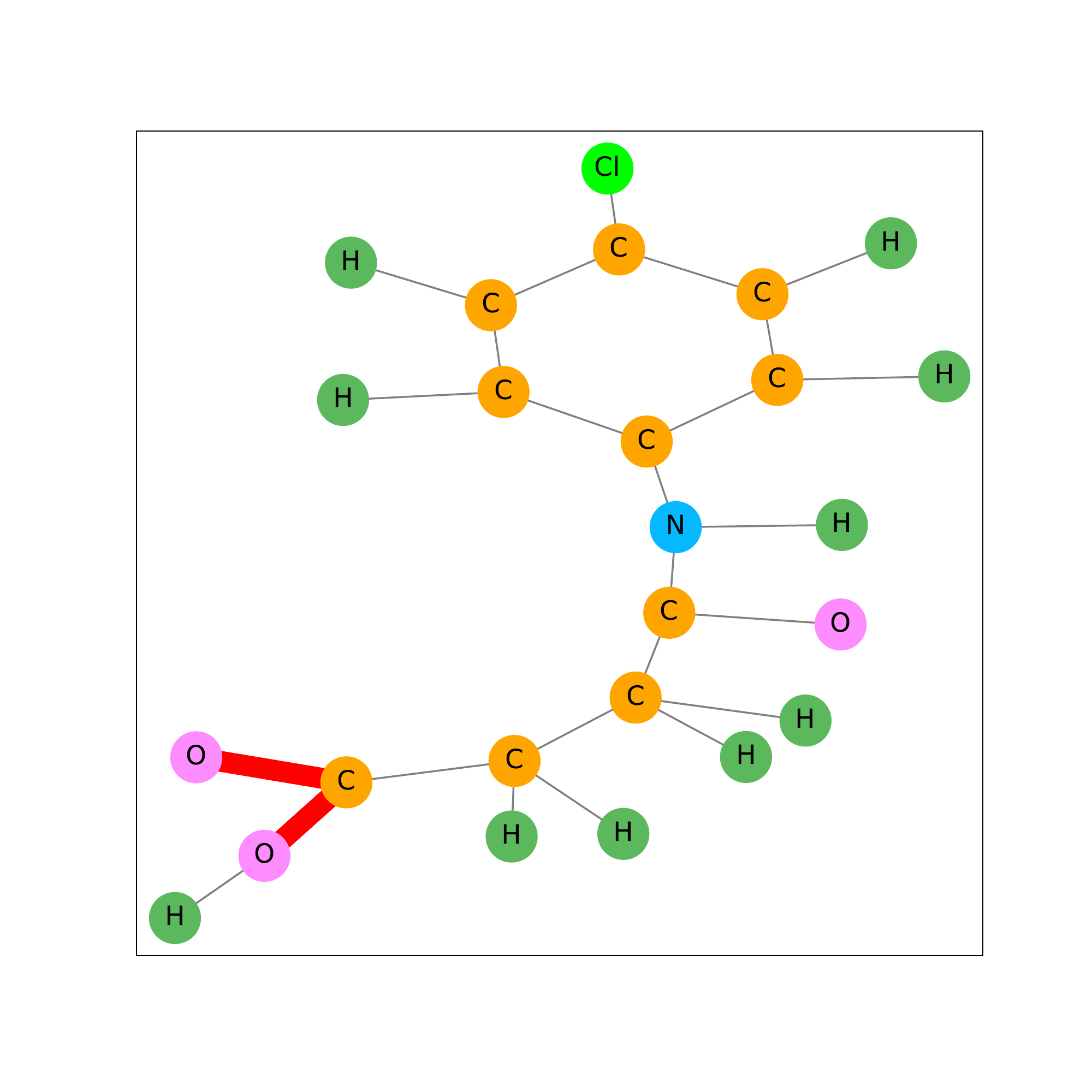}
    }
    \caption{A Structural Explanation of a graph in the Mutag dataset compared with explanations of Same-Class and Different-Class Samples. Red edges illustrate crucial edges.}
    \label{fig:graph_explanation}
\end{figure}

\noindent\textbf{Output.} \cref{fig:graph_explanation} presents structural explanations highlighting crucial edges for the predictions of a Mutag graph and reference samples. \cref{fig:graph_ex_wd} illustrates the web interface of these explanations. Highlighted edges first provide users with information about important motifs. Comparing explanations of samples in different classes gives them more insights into the target graph's prediction.

\section{Conclusion} \label{conclusion_part}
This paper presented INGREX, an interactive GNN explanation framework. The framework was developed based on state-of-the-art GNN explanation methods, advanced libraries, and tools. Specifically, INGREX can provide multilevel structural explanations and feature attributions for classification problems with graph data. Unlike existing GNN explanation methods, users can actively interact with explanations in our framework to better understand model decisions. We demonstrated INGREX in three explanation scenarios to present its effectiveness and helpfulness. Our framework is extensible and can be further improved to support a wide range of graph analytics problems.
\bibliographystyle{IEEEtran}
\bibliography{bibliography}

\end{document}